\documentclass[10pt]{article}
\usepackage{aaai}
\newcommand{\ul}{\underline}

\input{psfig}

\title{Lazy Transformation-Based Learning}
\author{Ken Samuel\\
	Department of Computer and Information Sciences\\
	University of Delaware\\
	Newark, Delaware 19716 USA\\
	samuel@cis.udel.edu\\
	http://www.eecis.udel.edu/\~{ }samuel/}

\begin{document}

\maketitle

\bibliographystyle{aaai}

\footnotetext[0]{Copyright \copyright\ 1998, American Association for Artificial
Intelligence (www.aaai.org). All rights reserved.}

\begin{abstract}
\begin{quote}

We introduce a significant improvement for a relatively new machine
learning method called Transformation-Based Learning. By applying a
Monte Carlo strategy to randomly sample from the space of rules,
rather than exhaustively analyzing all possible rules, we drastically
reduce the memory and time costs of the algorithm, without
compromising accuracy on unseen data. This enables
Transformation-Based Learning to apply to a wider range of domains, as
it can effectively consider a larger number of different features and
feature interactions in the data. In addition, the Monte Carlo
improvement decreases the labor demands on the human developer, who no
longer needs to develop a minimal set of rule templates to maintain
tractability.

\end{quote}
\end{abstract}

\section{Introduction}

Transformation-Based Learning (TBL)~\cite{Brill95a} is a promising new
machine learning algorithm, which has a number of advantages over
alternative approaches. However, one major limitation of TBL is that
it requires detailed information specifying the set of feature patterns that
are relevant to a particular problem. This imposes a significant
demand on the human developer. If he inadvertently omits any relevant
information, the learning process is handicapped, and, on the other
hand, if he includes too much information, the algorithm becomes
intractable, in practice. 

In this paper, we present a modification to TBL that enables the
algorithm to run efficiently, even when bombarded with an excessive
quantity of irrelevant information. This Lazy Transformation-Based
Learning (LTBL) method significantly reduces the demand on the
developer, who no longer needs to worry about excluding irrelevant
features from the input and must only insure that all of the
potentially relevant feature patterns are included. The key to our
solution involves using a Monte Carlo (random sampling) method. Unlike
the standard TBL method, which {\em exhaustively} searches for the
best model of the training data, LTBL only examines a relatively
small subset of the possibilities. Our experimental results show that
this modification drastically decreases the training time and memory
usage, without compromising the accuracy of the system on unseen data.

All of the examples and experimental results in this paper are drawn
from our work on a language understanding problem called Dialogue Act
Tagging, where the goal is to label each utterance in a conversational
dialogue with the correct {\em dialogue act}, which is an abstraction
of the speaker's intention~\cite{Samuel98a}. Examples of dialogue acts
are illustrated by the dialogue in Figure~\ref{ex-das}.

\begin{figure}[ht]
\centering
\begin{tabular}{|rlc|}
\hline
Speaker & Utterance & Dialogue Act \\
\hline
$\mathrm{A_{1}}$ & I have some problems                    & INFORM \\
                 & with the homework.                      & \\
$\mathrm{A_{2}}$ & Can I ask you a couple                  & REQUEST \\
                 & of questions?                           & \\
$\mathrm{B_{1}}$ & I can't help you now.                   & REJECT \\
$\mathrm{B_{2}}$ & Let's discuss it Friday...              & SUGGEST \\
$\mathrm{A_{3}}$ & Okay.                                   & ACCEPT \\
\hline
\end{tabular}
\caption{Dialogue between speakers A and B}
\label{ex-das}
\end{figure}

\section{Transformation-Based Learning}

TBL is a relatively new symbolic machine learning algorithm. When
tested on the Part-of-Speech Tagging problem,\footnote{The goal of
this task is to label words with part-of-speech tags, such as Noun and
Verb.} TBL was as effective as or better than the alternative
approaches, producing the correct tag for 97.2\% of the words in
unseen data~\cite{Brill95a}. In comparison with other machine learning
methods, TBL has a number of advantages, which we will present in a
later section.

\subsection{Labeling Data with Rules}

Given a training corpus, in which each entry is already labeled with
the correct tag, TBL produces a sequence of rules that serve as a
model of the training data. These rules can then be applied, in order,
to label untagged data.

\begin{figure}[ht]
\centerline{\psfig{figure=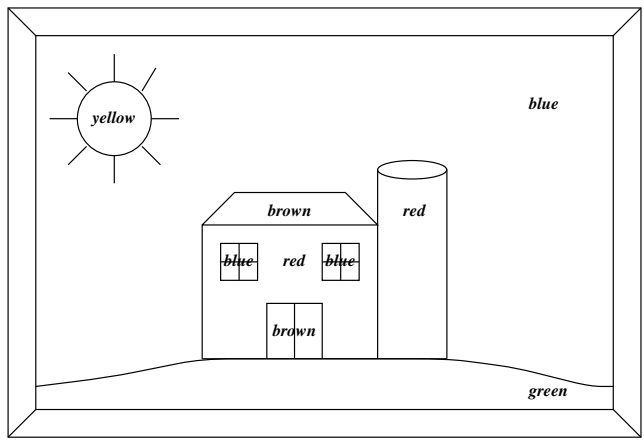}}
\caption{A barnyard scene} 
\label{tbl}
\end{figure}

The intuition behind the TBL method can best be conveyed by means of a
picture-painting analogy.\footnote{We thank Terry Harvey for
suggesting this analogy.} Suppose that an artist uses the following
method to paint a simple barnyard scene. (See Figure~\ref{tbl}.) He
chooses to begin with the blue paint, since that is the color of the
sky, which covers a majority of the painting. He takes a large brush,
and simply paints the entire canvas blue. After waiting for the paint
to dry, he decides to add a red barn. In painting the barn, he doesn't
need to be careful about avoiding the doors, roof, and windows, as he
will fix these regions in due time. Then, with the brown paint, he
uses a smaller, thinner brush, to paint the doors and roof of the barn
more precisely. He uses this same brush to paint green grass and a
yellow sun. Next, he returns to the blue to repaint the barn's
windows. And, finally, he takes a very thin, accurate brush, dips it
in the black paint, and draws in all of the lines.

The important thing to notice about this painting strategy is how the
artist begins with a very large, thick brush, which covers a majority
of the canvas, but also applies paint to many areas where it doesn't
belong. Then, he progresses to the very thin and precise brushes,
which don't put much paint on the picture, but don't make any
mistakes. TBL works in much the same way. The method generates a
sequence of rules to use in tagging data. The first rules in the
sequence are very general, making sweeping generalizations across the
data, and usually making several errors. Subsequently, more precise
rules are applied to fine-tune the results, correcting the errors, one
by one.

\begin{figure}[ht]
\centering
\begin{tabular}{|c|l|c|}
\hline
\# & \multicolumn{1}{c|}{Condition(s)} & New Dialogue Act \\
\hline
\hline
1 & {\em none}             & SUGGEST \\
\hline
2 & Change of Speaker      & REJECT \\
\hline
3 & Includes ``I''         & INFORM \\
\hline
4 & Includes ``Can''       & REQUEST \\
\hline
5 & Prev. Tag = REQUEST    & REJECT \\
  & Includes ``can't''     &        \\
\hline
6 & Current Tag = REJECT   & ACCEPT \\
  & Includes ``Okay''      &  \\
\hline
\end{tabular}
\caption{A sequence of rules}
\label{ex-rules}
\end{figure}

Figure~\ref{ex-rules} presents a sequence of rules that might be
produced by TBL for the Dialogue Act Tagging task. Suppose these rules
are applied to the dialogue in Figure~\ref{ex-das}. The first rule is
extremely general, labeling every utterance with the dialogue act,
SUGGEST. This correctly tags utterance B$_{2}$ in the sample dialogue,
but the labels assigned to the other utterances are not correct yet.
Next, the second rule says that, whenever a change of speaker occurs
(meaning that the speaker of an utterance is different from the
speaker of the preceding utterance), the REJECT tag should be applied.
This rule relabels utterances A$_{1}$, B$_{1}$, and A$_{3}$ with
REJECT. The third rule tags an utterance INFORM if it contains the
word, ``I'', which holds for utterances A$_{1}$, A$_{2}$, and B$_{1}$.
Next, the fourth rule changes the tag on utterance A$_{2}$ to REQUEST,
because it includes the word, ``Can''.

At this point, only utterances B$_{1}$ and A$_{3}$ are incorrectly
tagged. As we continue, the rules become more specific. The fifth rule
states that, if the previous tag (the tag on the utterance immediately
preceding the utterance under analysis) is REQUEST, and the current
utterance contains the word, ``can't'', then the tag of the current
utterance should be changed to REJECT. In the sample dialogue, this
rule applies to utterance B$_{1}$. And finally, the last rule changes
the tag on utterance A$_{3}$ to ACCEPT, so that all of the tags are
correct.

\subsection{Producing the Rules}

The training phase of TBL, in which the system learns a sequence of
rules based on a tagged training corpus, proceeds in the following
manner:

\smallskip

\begin{ttfamily}

\noindent
1.Label each instance with an initial tag.

\noindent
2.Until the stopping criterion is satisfied,\footnote{Typically,
the stopping criterion is to terminate training when no rule can be
found that improves the tagging accuracy on the training corpus by
more than some predetermined threshold~\cite{Brill95a}.}

a.For each instance that is currently 

\ \ tagged incorrectly,

\ i.Generate all rules that correct the tag.

b.Compute a score for each rule generated.\footnote{The score
measures the amount of improvement in the tagging accuracy of the
training corpus that would result from including a given rule in the
final model~\cite{Brill95a}.}

c.Output the highest scoring rule.

d.Apply this rule to the entire corpus.

\end{ttfamily}

\smallskip

This algorithm produces a sequence of rules, which are meant to be
applied in the order that they were generated. Naturally, some
restrictions must be imposed on the way in which the system may
compute rules for step 2ai, as there are an infinite number of rules
that can fix the tag of a given instance, most of which are completely
unrelated to the task at hand.\footnote{For example, the following
rule would correctly tag utterance B$_{2}$ in Figure~\ref{ex-das}:
{\em IF} the third letter in the second word of the utterance is
``s'', {\em THEN} change the utterance's tag to SUGGEST.} For this
reason, the human developer must provide the system with a set of {\em
rule templates}, to restrict the range of rules that may be
considered. Each rule template consists of a conjunction of zero or
more conditions that determine when a rule is applicable. Five sample
rule templates are illustrated in Figure~\ref{ex-templates}; these
templates are sufficiently general to produce all of the rules in
Figure~\ref{ex-rules}. For example, the last template can be
instantiated with \ul{X}=REQUEST, \ul{w}=``can't'', and \ul{Y}=REJECT
to produce the fifth rule.

\begin{figure}[ht]
\centering
\begin{tabular}{|rl|}
\hline
{\em IF}   & {\em no conditions} \\
{\em THEN} & change \ul{u}'s tag to \ul{Y} \\
\hline
{\em IF}   & \ul{u} includes \ul{w} \\
{\em THEN} & change \ul{u}'s tag to \ul{Y} \\
\hline
{\em IF}   & change of speaker for \ul{u} is \ul{B} \\
{\em THEN} & change \ul{u}'s tag to \ul{Y} \\
\hline
{\em IF}   & the tag on \ul{u} is \ul{X} \\
{\em AND}  & \ul{u} includes \ul{w} \\
{\em THEN} & change \ul{u}'s tag to \ul{Y} \\
\hline
{\em IF}   & the tag on the utterance preceding \ul{u} is \ul{X} \\
{\em AND}  & \ul{u} includes \ul{w} \\
{\em THEN} & change \ul{u}'s tag to \ul{Y} \\
\hline
\end{tabular}
\caption{A sample set of templates,
where \ul{u} is an utterance, \ul{w} is a word,
\ul{B} is a boolean value, and \ul{X} and \ul{Y} are dialogue acts}
\label{ex-templates}
\end{figure}

\section{Lazy Transformation-Based Learning}

Developing a workable set of rule templates is not a simple matter. If
the human developer inadvertently omits a relevant template from the
list, then the system cannot generate the corresponding rules, and so
its learning is handicapped. To increase the likelihood that all of
the relevant templates are available, the system should have access to
an overly-general set of rule templates.\footnote{In a later section,
we argue that TBL is capable of discarding irrelevant rules, so this
approach should be effective, in theory.} Unfortunately, if there are
too many templates, the TBL algorithm becomes intractable, because,
for each iteration, for each instance that is incorrectly tagged, {\em
every} template must be instantiated with the instance in {\em all}
possible ways. For some tasks, it might not even be {\em
theoretically} possible to capture all of the necessary information,
while still maintaining tractability.

Brill circumvented this problem by hand-selecting fewer than 30
templates, each consisting of only one or two
conditions~\cite{Brill95a}. Unfortunately, it is often very difficult
to construct such a limited set of templates without omitting any
relevant patterns. Satta and Henderson~\cite{Satta97} suggested an
alternative solution: They introduced a data structure that can
efficiently keep track of all possible transformations simultaneously,
which allows TBL to consider a large number of rule templates. But
their paper does not present any experimental results, and it is not
clear how effectively their method would work in practice.

In our work on applying TBL to Dialogue Act Tagging, we have developed
a modification to TBL, so that it may work efficiently and effectively
with a large number of templates. This LTBL method {\em randomly}
samples from the set of possible rules. In other words, for each
iteration and for each instance in the training set, only R rules are
generated, where R is some small integer.

Theoretically, for a given R, increasing the number of templates
should no longer affect the time and memory usage in training, since
the number of rules being considered for each iteration and each
instance is held constant. But even though only a small percentage of
the possible rules are actually being examined, we would expect LTBL
to continue to be successful when labeling unseen data, 
because the best rules are effective for several instances, so there
are several opportunities to find these rules. Thus, the better a rule
is, the more likely it is to be generated. And therefore, although
LTBL misses many rules, it is highly likely to find the best rules.

\section{Experimental Results}

\begin{figure}[t]
\centerline{\psfig{figure=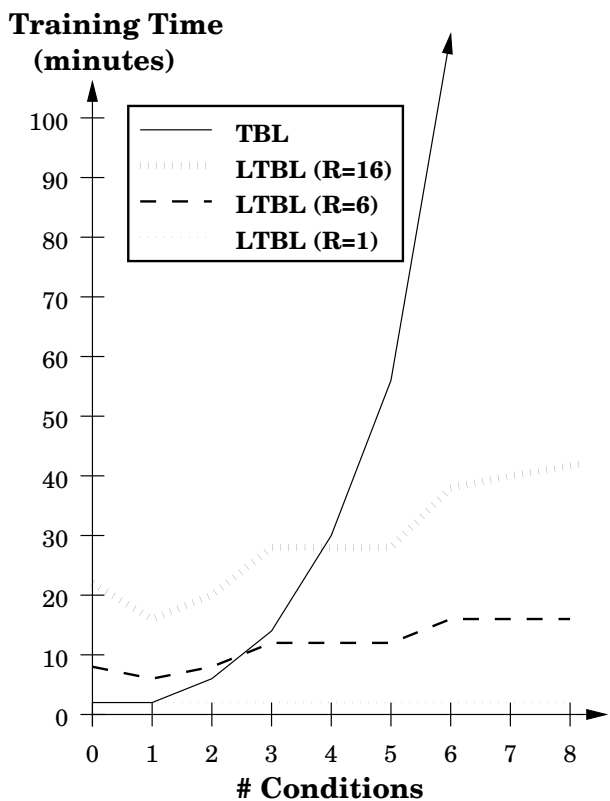}}
\caption{Number of conditions vs. training time}
\label{time-graph}
\end{figure}

\begin{figure}[t]
\centerline{\psfig{figure=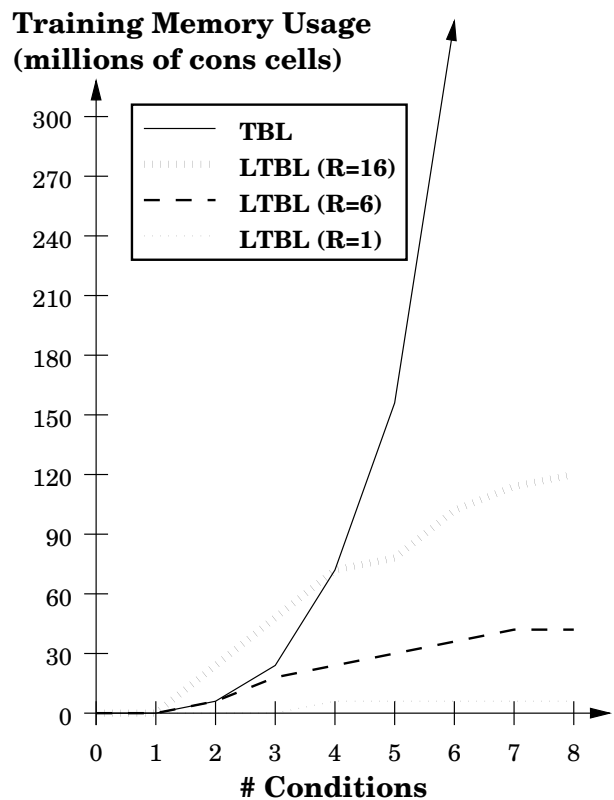}}
\caption{Number of conditions vs. training memory usage}
\label{memory-graph}
\end{figure}

\begin{figure}[t]
\centerline{\psfig{figure=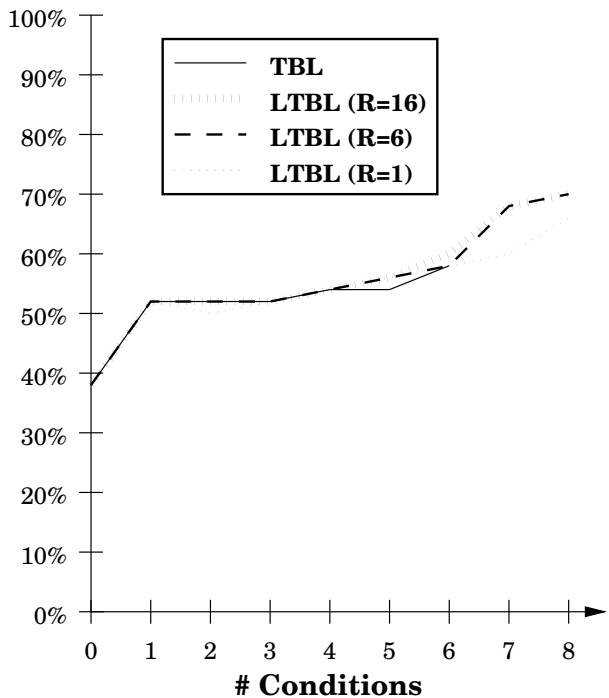}}
\caption{Number of conditions vs. accuracy in tagging unseen data}
\label{accuracy-graph}
\end{figure}

Some results from our Dialogue Act Tagging experiments are presented
in Figures~\ref{time-graph},~\ref{memory-graph},
and~\ref{accuracy-graph}. For these runs, a list of conditions was
preselected, and, for different values of n, $\mathrm{0 \leq n \leq
8}$, the first n conditions in the list were combined in all possible
ways to generate $\mathrm{2^n}$ possible templates. Using these
templates, we trained four methods on a training set and then
evaluated them with a disjoint testing set. We used a Sun Ultra 1 with
508MB of main memory for all of the experiments presented in this
paper.

Note that some conditions are more complex than others. For example,
the seventh condition in the list, which tests if a given utterance
contains specific patterns of words, generally had the greatest effect
on performance. But this increase in accuracy came with a price, as
templates with this condition tend to generate several rules for each
instance. In fact, when given the first seven conditions, the standard
TBL algorithm could not complete the training phase, even after
running for more than 24 hours.

Figures~\ref{time-graph} and~\ref{memory-graph} show that, for the
standard TBL method, time\footnote{The graph shows ``cpu time'', since
``real time'' is significantly influenced by unrelated factors.} and
memory usage rise dramatically as the number of conditions increases.
But LTBL keeps the efficiency relatively stable.\footnote{The reason
that LTBL may be slower than standard TBL in some cases is because
LTBL always generates R rules for each instance, without checking for
repetitions. (It would be too inefficient to prevent the system from
ever considering the same rule twice.)} (These curves increase
gradually, because, as the system is given access to more conditions,
it can discover a larger number of useful rules, resulting in more
iterations of the training algorithm before the stopping criterion is
satisfied.) By extrapolating the curves in Figure~\ref{time-graph}, we
would predict that LTBL with R=6 can train in under an hour with
trillions of templates, while the standard TBL method can only handle
about 32 templates in an hour.

Although these improvements in time and memory efficiency are very
impressive, they would be quite uninteresting if the performance of
the algorithm deteriorated significantly. But, as
Figure~\ref{accuracy-graph} shows, this is not the case. Although
setting R too low (such as R=1 for 7 and 8 conditions) can result in a
decrease in accuracy, LTBL with the lowest possible setting, R=1, is
as accurate as standard TBL for 64 templates.\footnote{One might
wonder how it is possible for LTBL to ever do better than the standard
TBL method, which occurs for 5 conditions. Because TBL is a greedy
algorithm, choosing the best available rule on each iteration,
sometimes the standard TBL method selects a rule that locks it into a
local maximum, while LTBL might fail to consider this attractive rule
and end up producing a better model.} The graph does not present
results for standard TBL with more than 6 conditions, because training
required too much time. But, as the curves for LTBL with R=6 and R=16
do not differ significantly, it is reasonable to conclude that
standard TBL would have produced similar results as well. Therefore,
LTBL (with R=6) works effectively for more than 250 templates in only
about 15 minutes of training time.

\subsection{Justifying the Use of TBL}

TBL has a number of advantages over other machine learning
methods.\footnote{Ramshaw and Marcus~\shortcite{Ramshaw94} presented
reasons for preferring TBL to Decision Trees.} An attractive
characteristic of TBL is its learned model: a relatively short
sequence of intuitive rules, stressing relevant features and
highlighting important relationships between features and tags. So,
TBL's output offers insights into a {\em theory} to explain the data.
This is a reason to prefer TBL over probabilistic machine learning
methods, since TBL's rules could ``allow developers to more easily
understand, manipulate, and debug the resulting
system.''~\cite{Brill97}

TBL is capable of discarding irrelevant rules, so it is not necessary
that all of the given rule templates be useful. If an irrelevant rule
is generated, its effect on the training corpus is essentially random,
resulting in a low score, on average. Thus, this rule is unlikely to
be selected for inclusion in the final model. Ramshaw and
Marcus~\shortcite{Ramshaw94} experimentally demonstrated TBL's
robustness with respect to irrelevant rules.

TBL is very flexible, in that it can accommodate many different types
of features, while other methods impose strong restrictions on their
features. Also, because of its iterative approach to generating rules,
TBL can utilize the tags that have been generated by the previous
rules as leverage for developing future rules. And, TBL can take
distant context into account with features that consider preceding
tags.

Since many machine learning methods may overfit to the training data
and then have difficulty generalizing to new data, they require that
additional measures be taken, such as cross-validation and pruning.
But Ramshaw and Marcus's~\shortcite{Ramshaw94} experiments suggest
that TBL tends to be resistant to this overtraining effect. This can
be explained by observing how the rule sequence produced by TBL
progresses from general rules to specific rules. The early rules in
the sequence are based on many examples in the training corpus, and so
they are likely to generalize effectively to new data. And, later in
the sequence, the rules don't receive much support from the training
data, and their applicability conditions tend to be very specific, so
they have little or no effect on new data. Thus, resistance to
overtraining is an emergent property of the TBL algorithm.

\section{Summary}

Current implementations of TBL break down, in practice, with a very
limited number of templates. This research provides a solution that
can work efficiently with hundreds of templates, without suffering a
decrease in accuracy, thereby increasing the applicability of this
promising machine learning method and lessening the labor demands on
the human developer. Our experiments suggest that, for 250 templates,
LTBL can train in about fifteen minutes, while the standard TBL method
would require days of training time to produce comparable results.

\section{Acknowledgments}

I wish to thank Sandra Carberry and K. Vijay-Shanker, for their
suggestions on this paper.

I also owe thanks to the members of the \scshape VerbMobil \normalfont
research group at DFKI in Germany, particularly Norbert Reithinger,
Jan Alexandersson, and Elisabeth Maier, for providing me with the
opportunity to work with them and generously granting me access to the
\scshape VerbMobil \normalfont corpora so that I could test my system.

This work was partially supported by the NSF Grant \#GER-9354869.

\bibliography{/usa/samuel/class/research/related_research/mybibfile}

\end{document}